\documentclass[lettersize,journal]{IEEEtran}
\usepackage{amsmath,amsfonts}
\usepackage{algorithmic}
\usepackage{algorithm}
\usepackage{array}
\usepackage[caption=false,font=normalsize,labelfont=sf,textfont=sf]{subfig}
\usepackage{textcomp}
\usepackage{stfloats}
\usepackage{url}
\usepackage{verbatim}
\usepackage{graphicx}
\usepackage{cite}
\usepackage{xcolor}

\usepackage[export]{adjustbox}
\usepackage{multirow}

\begin{document}

\title{PLGAN: Generative Adversarial Networks for Power-Line Segmentation in Aerial Images}

\author{Rabab Abdelfattah, Xiaofeng Wang,~\IEEEmembership{Member,~IEEE,} Song Wang,~\IEEEmembership{Senior Member,~IEEE}
\thanks{Manuscript received December 31, 2021. This work was supported by the National Science Foundation (ECCS1830512). (\emph{Corresponding authors: Xiaofeng Wang; Song Wang}.)

Rabab Abdelfattah and Xiaofeng Wang are with Department of Electrical Engineering and Computing, University of South Carolina, Columbia, SC 29208 USA (e-mail: rabab@email.sc.edu; wangxi@cec.sc.edu).

Song Wang with the Department of Computer Science and Engineering, University of
South Carolina, Columbia, SC 29208 USA (e-mail: songwang@cec.sc.edu).
}
}

\markboth{}
{Shell \MakeLowercase{\textit{et al.}}:}


\maketitle

\begin{abstract}

Accurate segmentation of power lines in various aerial images is very important for UAV flight safety. The complex background and very thin structures of power lines, however, make it an inherently difficult task in computer vision. This paper presents PLGAN, a simple yet effective method based on generative adversarial networks, to segment power lines from aerial images with different backgrounds. Instead of directly using the adversarial networks to generate the segmentation, we take their certain decoding features and embed them into another semantic segmentation network by considering more context, geometry, and appearance information of power lines. We further exploit the appropriate form of the generated images for high-quality feature embedding and define a new loss function in the Hough-transform parameter space to enhance the segmentation of very thin power lines. Extensive experiments and comprehensive analysis demonstrate that our proposed PLGAN outperforms the prior state-of-the-art methods for semantic segmentation and line detection.

\end{abstract}

\begin{IEEEkeywords}
Power-line segmentation, generative adversarial networks, image segmentation, aerial images, line detection.
\end{IEEEkeywords}

\vskip 0pt 
\section{Introduction}
\label{sec:int}
While unmanned aerial vehicles (UAVs) have been used in many recreational, photography, commercial and military applications, their flight safety may be threaten by the widespread power lines~(PLs)~\cite{bhanu1996system}.  Hitting a PL may not only destruct the UAVs but also damage power grids and electrical properties as well. Given their very thin structures, however, PLs are prone to be missed by many detection sensors. To enable UAVs to detect and localize PLs during flight, this paper presents a new computer-vision approach aiming to accurately segment PLs from \textit{aerial images} that are taken by the cameras mounted on UAVs.

PL segmentation from aerial images is very challenging. From a bird's-eye view, the background of aerial images can be any places, e.g., desert, lakes, mountains, and cities, which shows significant variety and complexity. Moreover, PLs and their surrounding background may share very similar color in many cases and therefore are difficult to distinguish from local image information.  Finally, PLs have very thin structures and only cover a very small portion of the image, e.g., one- or few-pixel wide in aerial images. As a result, the PL segmentation is vulnerable to be fragmented, leading to poor segmentation performance.

There have been many deep-learning based algorithms developed for achieving state-of-the-art performance on general-purpose line-segment detection~\cite{huang2018learning,xue2020holistically,zhang2019ppgnet,zhou2019end}, most of which rely on the saliency of lines and joint inference of junctions.  Both of these properties, however, do not hold for PLs in most aerial images. The recent AFM model~\cite{xue2019learning} detects line segments by constructing an attraction field map instead of inferring junctions. Nevertheless, it cannot handle well the background complexity in aerial images, as verified in our later experiments.
PL segmentation can be treated as a kind of semantic image segmentation, for which many advanced deep neural networks, such as FCN~\cite{long2015fully} and DeepLab~\cite{chen2017rethinking,chen2018encoder}, have been developed with state-of-the-art performance on public image dataset, such as Cityscape and PASCAL VOC. However, without considering the shape and inter-pixel relations, these semantic segmentation networks cannot accurately capture very thin PLs with similar color to the surrounding background in aerial images.

To find the inter-pixel relations and enforce the global consistency between pixels, in this paper we propose to use generative adversarial networks (GANs) as a backbone for PL segmentation. The main motivation is to leverage the min-max loss of GANs to help 1) generate a natural (real) image with good reflection of the relationship between adjacent pixels, and 2) create a high-quality feature embedding for semantic image segmentation. Specifically, this paper presents a new PLGAN (PL Generative Adversarial Networks) to segment PLs from aerial images by employing adversarial learning. In the proposed PLGAN,  we first include a multi-task encoder-decoder network to generate an image with highlighted PLs. Then we take the last feature representation (i.e., the one right before the output layer) of the decoder network and embed it to a semantic segmentation network to improve PL segmentation. We define comprehensive loss functions, including adversarial, geometry, and cross entropy ones, for PLGAN training. Furthermore, we include a loss function in the Hough transform parameter space to highlight the long-thin nature of PLs. Extensive experiments, including ablation studies and comparison experiments with prior methods, on the public TTPLA dataset~\cite{abdelfattah2020ttpla} and Massachusetts roads dataset for road segmentation \cite{mnih2013machine}, verify the effectiveness of the proposed method.

Our main contributions are summarized as below.
\begin{itemize}
 \item 
 \vskip 0pt 
A novel PLGAN network is proposed to segment very-thin PLs from aerial images with complex backgrounds. To the best of our knowledge, this is the first generative adversarial network~(GAN) developed for line structure segmentation. The novelty comes from using the PL-highlighted images for discrimination and introducing the semantic decoder with the most representative embedding vectors as the input to generate semantic images.
\item
 \vskip 0pt
 A new loss function is introduced in the \textbf{modified} Hough transform parameter space and then combined to adversarial, geometry, and cross entropy losses to enhance PL segmentation performance.
\item
 \vskip 0pt 
The proposed PLGAN significantly promotes the state-of-the-art performance of PL segmentation on the public TTPLA dataset. To show the generality of the proposed method, we also apply PLGAN to the Massachusetts Roads dataset for segmenting roads, which are also long and thin as PLs. 
\end{itemize}

For the remainder of the paper, Section~\ref{sec:related} briefly overviews the related work. Section~\ref{sec:approach} elaborates on describing the proposed PLGAN. Section~\ref{sec:results} reports the experimental results, followed by a brief conclusion in Section~\ref{sec:con}.

\section{Related Work}
\label{sec:related}

The related work is discussed in four parts: power lines~(PLs), line segment detection, semantic segmentation, and GANs.


\subsection{Power Lines}
\label{sec:pl-related}
Most existing PL-related datasets were managed to own nice properties to simplify PL detection, such as synthetic PLs~\cite{jenssen2021ls}, manually cropping aerial images to obtain subimages focused on PLs~\cite{zhang2019detecting}, and capturing images from ground~\cite{russakovsky2015imagenet}, to name a few. Compared with these datasets, TTPLA dataset we use in this paper is more challenging and practical.  It includes aerial images with very complex background and wide varieties in zoom levels, view angles, time during a day, as well as weather conditions~\cite{abdelfattah2020ttpla}.

Most existing work on PL detection adopts traditional computer vision methods~\cite{candamo2009detection, yan2007automatic, golightly2005visual, li2008knowledge, santos2017plined}, which have multiple drawbacks. First, it is often assumed that the PLs are parallel and straight so that context-assisted information can be used to extract PLs~\cite{yan2007automatic,li2008knowledge}, while this assumption may not hold in practice. Second, extracting edge maps with traditional approaches requires good contrast between the PLs and the surrounding background which can only be achieved in ideal cases~\cite{yetgin2017comparison}.  In practice, the color of the PLs and the background could be very similar in aerial images. 
Third, traditional methods usually rely on predefined hyper-parameters to generate meaningful results. However, defining these hyper-parameters is very challenging, especially for those datasets with images taken in a wide range of conditions (e.g., different zoom levels, points of view, background, light, and contrast).  

Recently, deep-learning based methods were investigated~\cite{zhang2019detecting, yetgin2018power, jenssen2021ls,li2019attentional, madaan2017wire, lee2017weakly, zhao2019region} for PL detection.
Yetgin et al. \cite{yetgin2018power} proposed an end-to-end CNN architecture with randomly initialized softmax layer for jointly fine-tuning the feature extraction and binary classification -- PL and non-PL background are classified at the image level. Yetgin et al. further developed a feature classification method for PL segmentation, where features are extracted from the intermediate stages of the CNN. In~\cite{pan2016power} a CNN-based classifier is developed to identify the input-image patch that contains PL and then use Hough transform as the post-processing  to localize the PLs in each patch. In~\cite{gubbi2017new},  a deep CNN architecture with fully connected layers is proposed for PL segmentation, where the CNN inputs are histogram-of-gradient features -- a sliding window is moved over each patch to get a classification of PL or not.
In \cite{jaffari2021novel}, a UNET architecture is trained to segment PLs based on a generalized focal loss function that uses the Matthews correlation coefficient~\cite{matthews1975comparison} to address the class imbalance problem. In~\cite{li2019attentional}, an attentional convolutional network is proposed for pixel-level PL detection, and it consists of an encoder–decoder information fusion module and an attention module, where the former fuses the semantic information and the location information while the latter focuses on PLs. In~\cite{madaan2017wire}, dilated conventional networks with different architectures are tried by finding the best architecture over a finite space of model parameters. Choi et al.~\cite{lee2017weakly} proposed a weakly supervised learning network for pixel-level PL detection using only image-level classification labels. However, besides the simplicity of the datasets as mentioned before, most of these CNN-based works formulate the problem as pixel-wise classification with convolutional neural networks (CNNs) and do not well consider global consistency in detection, which is essential in detecting very thin structures~\cite{zhao2019region}.   
\subsection{Line Segment Detection}
Significant progress has been made on line segment detection in recent years 
by using deep neural networks.
Most of deep line detection approaches rely on the junction information to locate the true line segments: Some jointly detect the junctions and line segments~\cite{huang2018learning,xue2020holistically}, while many others detect only the junctions and then use sampling methods to deduce the line segments~\cite{zhang2019ppgnet,zhou2019end}. 
These methods are not applicable to our task since PLs in aerial images may not always be straight and usually lack junctions. 

\subsection{Semantic Segmentation}
Deep neural networks for semantic segmentation~\cite{chen2014semantic,zheng2015conditional} rely on pooling layers to reduce the spatial resolution in the deepest FCN layers. As a result, prediction around the segmentation boundaries are usually poor due to insufficient contextual information~\cite{zhao2019region, ke2018adaptive, bertasius2017convolutional, huang2019ccnet}. Dilated convolutions are imported to capture larger contextual information \cite{chen2014semantic,YuKoltun2016,chen2017rethinking, chen2017deeplab, ding2018context}, which,  however, still cannot generate global context just from a few neighboring pixels~\cite{bertasius2017convolutional}. The encoder-decoder structures are emerged to overcome the drawback of atrous convolutions~\cite{ronneberger2015u,badrinarayanan2017segnet,lin2017refinenet}. However, the prediction accuracy is still limited when recovered from the fused features~\cite{tian2019decoders}. In addition, the softmax cross entropy loss limits semantic segmentation performance~\cite{ke2018adaptive,zhao2019region} by ignoring the correlation between pixels.
Many of these limitations can be maximized in segmenting very thin PLs and we will include several of the above methods into our comparison experiments.

\begin{figure*}[th]
	\centering
		\includegraphics[width=1\textwidth]{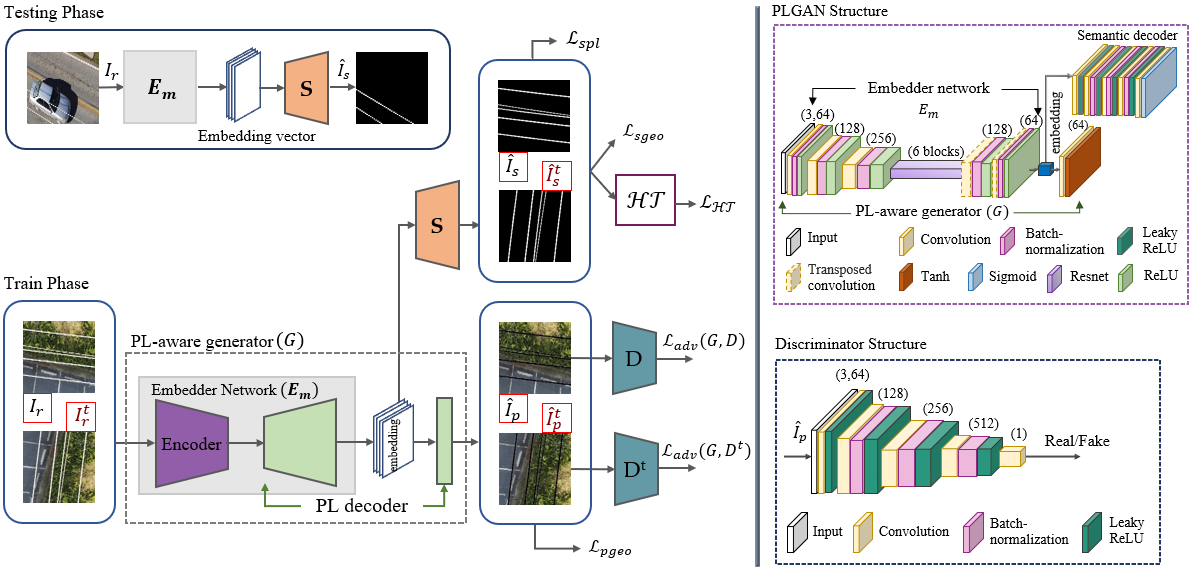}
			\caption{An illustration of PLGAN framework. PLGAN consists of the PL-aware generator $G$, two discriminators $D$ and $D^t$, and the semantic decoder $S$.  The PL-aware generator contains the encoder and PL-decoder. The embedder network $E_m$, included in the generator, consists of the encoder and the PL-decoder except the last output layer. The input to PLGAN is RGB image $I_r$ and its transformed input image $I_r^t$ (They are applied individually, not at the same time). The output of PLGAN is the semantic image for PLs $\hat I_s$. The $G$, $S$, $D$, and $D^t$ forms adversarial training to generate PL-highlighted image $\hat I_p$ from $G$ and the embedding vector from $E_m$. The embedding vector, which carries the context, appearance, and geometry information, is used as the input to $S$. The generated PL-highlighted images $\hat I_p$ and the transformed $\hat I_p^t$ are the inputs to $D$ and $D^t$, respectively. The PL-aware generator and the semantic decoder are jointly trained by the combination of adversarial, semantic, geometry, and Hough transform loss functions. There is no overhead during testing, only $E_m$ and $S$ networks are used to generate semantic images.}
			\vspace{-3mm}
	\label{fig:gan_framework}
\end{figure*}

\subsection{GANs} \label{subsec:GAN}
\label{sub:gan}
Generative Adversarial Networks (GANs)~\cite{goodfellow2014generative} has been widely used in image translation~\cite{isola2017image,zhu2017unpaired}, super-resolution~\cite{lai2017deep}, inpainting~\cite{pathak2016context}, and image editing/manipulation~\cite{zhu2016generative}.
However, directly applying GANs for segmentation may not be desirable for two reasons. First, GANs usually employ softmax loss at the output layer and this will prevent the networks from expressing uncertainties when generating semantic images~\cite{samson2019bet}. Second, the softmax probability vectors cannot produce exact zeros/ones, while the discriminator requires sharp zeros/ones. As a result, the discriminator may examine the small, but always existing, value gap between the distributions of the fake and real samples and needlessly learn more complicated geometrical discrepancies. In this paper, we embed GAN-extracted features for enhancing PL segmentation, instead of directly discriminating the semantic images.

\section{PLGAN Approach }
\label{sec:approach}
\noindent
\textbf{Notations.} Let $I_r \in \mathbb{R}^{w \times h \times c}$ denote the input image, where $w \times h$ is the dimensions of the input image and $c$ is the number of channels.  Let $\hat I_s \in \mathbb{R}^{w \times h }$ be the semantic output of PLGAN as shown in Figure~\ref{fig:gan_framework} and $\hat I_p \in \mathbb{R}^{w \times h \times c}$ be the PL-highlighted image (or ``fake image'') projected from the embedding vector $E_m(I_r)$. For the ground-truth of PL-highlighted image, we simply set the intensity of PL pixels in an image to zero. $I_s$ and $I_p$ are the ground truth (GT) of $\hat I_s$ and $\hat I_p$, respectively. Let $\phi:\mathbb{R}^{w \times h \times c} \to \mathbb{R}^{h \times w \times c}$ denote the geometry transformation on an image. Given an image $I$, the transformed image is denoted as $I^t = \phi(I)$. To make sure that the transformed images have the appropriate dimensions as the inputs to PLGAN, we assume $w = h$ (otherwise, zeros can be filled to ensure this equivalence).  Given a matrix $P$, $[P]_{ij}$ denotes the entry at the $i$th row and the $j$th column of $P$.  Accordingly, given an image $I$, $[I]_{ij}$ denotes the value at pixel $(i,j)$ in the image.  Given two cascaded functions or networks $G$ and $\phi$, $G\circ \phi(\cdot) = G(\phi(\cdot))$. $\|\cdot \|_1$ is the $L_1$ norm to calculate the absolute difference on each pixel. 

\subsection{PLGAN Structure $(G_{PL})$}
Our objective is to develop a deep neural network that predicts the semantic image $\hat I_s$ based on the input image $I_r$. The proposed PLGAN structure consists of the PL-aware generator, two discriminators, and the semantic decoder.  The discriminators are trained in an adversarial way against the PL-aware generator and the semantic decoder.
As shown in Figure~\ref{fig:gan_framework}, the embedder network $E_m$ maps the input image $I_r$ to the latent space where the generated embedding vector $E_m(I_r)$ carries the context, appearance, and geometry information.  This vector is mapped back to the image space through the output layer of the PL decoder and the semantic decoder for the PL-highlighted image $\hat I_p$ and the semantic image $\hat I_s$, respectively. During training, PLGAN will learn the features of the PL pixels and distinguish them from the background pixels based on adversarial loss functions. During testing, there is no additional overhead or post-processing steps, only $E_m$ and $S$ networks are used to generate semantic images.

The \textbf{PL-aware generator ($G$)} consists of the encoder and the PL decoder with the residual blocks~\cite{he2016deep} in the middle. The encoder and the PL decoder are composed of a sequence convolution layers and transpose convolutional layers with a stride of 2, respectively, both followed by batch-normalization and ReLU activation, as shown in Figure~\ref{fig:gan_framework}. 

The \textbf{semantic decoder ($S$)} outputs semantic images $\hat I_s$. In PLGAN, the embedding vector is directly sent to this decoder without going to the discriminator.  Instead, the discriminator focuses on the PL-highlighted images and its GT, which are color images.  By doing so, the benefits of adversarial training can be fully explored, which, as discussed in Subsection~\ref{subsec:GAN}, cannot be achieved by directly applying GANs (PL-aware generator and discriminator only). The semantic decoder consists of a set of convolution, batch-normalization, leaky-ReLU layers, and a sigmoid nonlinear layer as the output layer, as shown in Figure~\ref{fig:gan_framework}.

The \textbf{adversarial discriminator ($D$)} is to distinguish the PL-highlighted image $\hat I_p$ from its GT $I_p$.  Notice that $\hat I_p$ is very similar to the input image $I_r$ except that the PL pixels are highlighted. The PL area in $\hat I_p$ has a high-frequency structure because of sharp changes in intensity from the background pixels to the highlighted pixels.  Given this high-frequency nature, Markovian discriminator structure is used for its efficiency in tracking high-frequency structures~\cite{isola2017image}. The Markovian discriminator maps $\hat I_p$ at the patch level (i.e., patches are individually quantified to the fake or real value) and considers the structural loss, such as structural similarity, feature matching, and conditional random field, which will help compensate the loss of $\hat I_p$ at low frequencies. With these benefits, the discriminator is able to push the PL-aware generator to create more natural PL-highlighted images~\cite{lee2018deep}. Besides $D$, an additional discriminator $(D^t)$ is added to discriminate the transformed PL-highlighted image $\hat I_p^t$ and the GT $I_p^t$, which has a similar structure to $D$ as shown in Figure~\ref{fig:gan_framework}. 
\subsection{Objective Formulation }
The loss functions for different modules in PLGAN are defined as follows. 

\smallskip
\noindent
\textbf{Adversarial Loss.} The adversarial loss is applied to encourage $G$ to fool the discriminator $D$ by generating images that looks similar to the real images. While, $D$ is trained to distinguish between the real images $(I_p)$ and fake images $(\hat I_p)$. The least square loss function is chosen for our train, instead of  binary cross-entropy~\cite{fu2019geometry}, for more stable training and convergence~\cite{mao2017least}. The adversarial loss is defined~as:
\begin{align}
\label{eq:adv}
&\mathcal{L}_{adv}(G,D;I_r,I_p) \\
&=\frac {1}{2} \mathbb{E}_{I_p} \left[ (D(I_p))^2 \right]+\frac {1}{2} \mathbb{E}_{I_r} \left[(1-D\circ G(I_r))^2\right] \nonumber
\end{align}
where $\mathbb{E}_{I_p}$ and $\mathbb{E}_{I_r}$ are the empirical estimated expectations.   The discriminator $D$ is to maximize $\mathcal{L}_{adv}$ and $G$ is to minimize this loss, which formulates adversarial training. 

\smallskip
\noindent
\textbf{Semantic Loss.}  The cross entropy loss between $I_s$ and $\hat I_s$ is defined as follows: 
\begin{align} 
    & \mathcal{L}_{spl}(E_m,S;I_r,I_s) =\label{eq:sem}
    \\&  \frac{\sum_{(i,j) \in \mathcal N} \left([I_s]_{ij} \log([\hat I_s]_{ij}) +  (1-[I_s]_{ij}) \log(1-[\hat I_s]_{ij})\right)}{-|\mathcal N|} \nonumber 
\end{align}
where $\mathcal N$ is the pixel set of interest (e.g., pixels belonging to PLs), $|\mathcal N|$ is the number of elements in $\mathcal N$, and $\hat I_s = S\circ E_m(I_r)$. 
It is worth mentioning that the semantic loss is determined pixel by pixel, which may not be able to capture the correlations between pixels. In this case, missing one PL pixel may lead to spatially-disjoint object segments, given that PLs are very thin in aerial images (e.g., 1 pixel width). To address this issue, we introduce Hough transform loss function, which will be discussed next.

\smallskip
\noindent
\textbf{Hough Transform Loss.}
The motivation of using Hough transform loss is to force PLGAN to find and correct the flawed pixels along PLs so as to ensure global consistency for each PL. Each pixel in the semantic image is mapped to a sinusoidal curve in the parametric space by the modified Hough transform
\begin{equation}
\label{eq:ht}
 \mathcal{HT}([I_s]_{ij})= p_{ij} (i \cos \theta + j \sin \theta)
\end{equation}
where $(i,j)$ is the pixel coordinate in the semantic image $I_s$, $p_{ij} \in [0,1]$ is the confidence score at pixel $(i,j)$, $\theta \in [0,\theta_{\max})$ is the angle parameter, and $\theta_{\max}$ is the maximum value of $\theta$ (e.g., $\theta_{\max} = \pi$). During training, $p_{ij}$ will eventually approach to the neighborhood of 1 or 0, indicating that $(i,j)$ belongs to the PLs or not, respectively.  Otherwise, it will lead to a large of loss in the parameter space and force PLGAN to refine its prediction.
%
%
In practice, we partition the set $[0,\theta_{\max})$ into $M$ pieces and therefore one pixel in segmentation-image space will result in $M$ outputs, $y_{ij}(\theta_l)$, in parameter space, where $\theta_l = \frac{l\theta_{\max}}{M}$ for $l=0,1,\cdots,M-1$.  Missing one pixel in segmentation-image space implies missing $M$ points in parameter space.  As a result, the penalty will be magnified $M$ times in parameter space, which will make PLGAN to correct the flawed pixels. The Hough transform Loss is defined as follows.
\begin{align}
&\mathcal{L}_{\mathcal{HT}}(E_m,S;I_r,I_s)=  \mathbb{E}_{I_r} \left[\| {\mathcal{HT} (I_s) - \mathcal{HT} (\hat I_s)}\|_1\right]
\end{align}
with $\hat I_s= S\circ E_m(I_r)$, where $[\mathcal{HT} (I_s)]_{ij} = \mathcal{HT} ([I_s]_{ij})$.
%
%
From the other point of view, the pixels belong to the same PL in segmentation-image space are intersected as sinusoidal curves into the parameter space and accumulated as a value into the same cell into the discrete parametric space. Therefore, the intersection points have strong intensities as a results of intersecting more than one curve into the same point in the parameter space. Each intersection point includes two specifics: --The intersection point represents multiple related pixels belong to the same PL into the segmentation-image space. -- If the intensity of the intersection point is reduced as a result of missing one or more curves in the parameter space, the network should be penalized to learn finding the missing curves. Hence, the missing pixels in the segmentation-image space are recovered. Consequently, our Hough transform loss function guarantees global consistency for the PLs in the segmentation-image~space. 

\smallskip
\noindent
\textbf{Geometry Loss.}   According to~\cite{abdelfattah2020ttpla}, the PLs in average take 1.6\% of the total pixels in an aerial image.  In addition, the color of PLs in aerial images may be close to the background. Both facts indicate that the visual evidence of PLs is very weak. There is a possibility of generating trivial $\hat I_p$ that is very similar to the background in colors and styles while removing or decreasing the foreground. The discriminator may not be able to correctly identify the flawed pixels in $\hat I_p$ from the GT $I_p$ due to high similarity between $\hat I_p$ and $I_p$ at most pixels.  To address this issue, we add penalties on geometry space that force the training to correct failures in the local regions of PLs after geometry transformation.
Inspired by GcGAN~\cite{fu2019geometry}, geometry consistency focuses on the PL-highlighted image $\hat I_p=G(I_r)$ and the inverse of its transformed image $\phi^{-1}\circ G \circ \phi(I_r)$.  In PLGAN, we also consider geometry consistency between the semantic image $\hat I_s = S\circ E_m(I_r)$ and the inverse of its transformed semantic image $\phi^{-1}\circ S\circ E_m \circ \phi(I_r)$.   
The geometry loss is defined as follows:
\begin{align*}
&    \mathcal{L}_{pgeo}(G;I_r)= \mathbb{E}_{I_r} \left[\| G(I_r) - \phi^{-1}\circ G\circ \phi(I_r)\|_1\right] \\
    &~~~~~~~~~~~~~~ + \mathbb{E}_{I_r} \left[\| G\circ \phi(I_r) - \phi\circ G(I_r)\|_1\right] \\
&\mathcal{L}_{sgeo}(E_m,S;I_r) \\& ~~~~~~~~~~~~ =  \mathbb{E}_{I_r} \left[\| S\circ E_m(I_r) -\phi^{-1}\circ S \circ E_m \circ \phi(I_r)\|_1\right]\\
    &~~~~~~~~~~~~~~~~ + \mathbb{E}_{I_r} \left[\| S\circ E_m \circ \phi(I_r) - \phi\circ S\circ E_m(I_r)\|_1\right].
\end{align*}
With the penalty on the geometry loss, it is unlikely that $G$ and $G \circ \phi$ both fail at the same location. Instead, they co-regularize each other to keep geometry-consistency~\cite{fu2019geometry}.  So do $S \circ E_m$ and $S\circ E_m \circ \phi$.  Similarly, we can define the adversarial loss, the semantic loss, and the Hough transform loss in the transformed domain as $\mathcal{L}_{adv}(G,D^t;I_r^t,I_p^t)$, $\mathcal{L}_{spl}(E_m,S;I_r^t,I_s^t)$, and $\mathcal{L}_{\mathcal{HT}}(E_m,S;I_r^t,I_s^t)$, respectively, with $I_r^t = \phi(I_r)$, $I^t_s = \phi(I_s)$, and $I^t_p = \phi(I_p)$. $D^t$ is the discriminator for the transformed generated image $I_p^t$.

{The \textbf{overall loss} function can be defined as follows:}
\begin{align}
\label{eq:full2}
& \mathcal{L}_{G_{PL}}(G,D,D^t,S;I_r,I_s,I_p) \\
= &~ \mathcal{L}_{adv}(G,D;I_r,I_p) + \mathcal{L}_{adv}(G,D^t;I_r^t,I_p^t) \nonumber\\ 
  & + \lambda_{spl} \left( \mathcal{L}_{spl}(E_m,S;I_r,I_s) + \mathcal{L}_{spl}(E_m,S;I_r^t,I_s^t)\right) \nonumber\\  
  & + \lambda_{\mathcal{HT}} \left(\mathcal{L}_{\mathcal{HT}}(E_m,S;I_r,I_s) + \mathcal{L}_{\mathcal{HT}}(E_m,S;I_r^t,I_s^t) \right) \nonumber\\
  & + \lambda_{geo}\left(\mathcal{L}_{pgeo}(G;I_r)+\mathcal{L}_{sgeo}(E_m,S;I_r)\right) \nonumber
\end{align}
\begin{table*}[!ht]
    \centering
\caption{Quantitative PL segmentation performance of the proposed PLGAN and the comparison methods on TTPLA dataset~\cite{abdelfattah2020ttpla}. Bold represents the highest results and underline represents the second-best.}
\label{tab:results-basics}
\begin{adjustbox}{width=0.98\textwidth} 
    \begin{tabular}{l|c|c|c|c|c|c|c|c|c|c}
    \hline
        Models &Backbone&\multicolumn{1}{c|}{Percsion}&\multicolumn{1}{c|}{Recall}&\multicolumn{1}{c|}{IoU}&\multicolumn{1}{c|}{$F_1$}&\multicolumn{1}{c|}{$F_{\beta}$} & Corr & Comp & Quality & param (M) $\downarrow$ \\ 
        \hline
        FPN~\cite{lin2017feature} &Resnet$-34$& 0.769 & 0.513 & 0.423 & 0.569 & 0.635  & 0.884 & 0.743 & 0.674& 23.2   \\ 
        UNET\ &Resnet-34& 0.846 & 0.583 & 0.515& 0.662 & 0.735 & 0.904 & 0.823 & 0.754& 24.4   \\ 
                LinkNet \cite{chaurasia2017linknet}\ &Resnet-34& 0.836 & 0.569 & 0.496& 0.645 & 0.719 & 0.903 & 0.809 & 0.741 & 21.8  \\ 
        UNET++ \cite{zhou2018unet++}\ &Resnet-34& 0.843 & \textbf{0.591} & \underline{0.522}& \underline{0.668} & \underline{0.739} & 0.896 & \underline{0.833} & 0.760 & 26.1  \\

        MaNet~\cite{fan2020ma}\ &Resnet-34& \underline{0.858} & 0.585 & 0.517& 0.663 & 0.738 & \textbf{0.923} & 0.810 & \underline{0.759} & 31.8 \\ 
        FPN~\cite{lin2017feature} &Resnet-18&0.746&0.492&0.401&0.546&0.612&0.867&0.717&0.646&13.0\\
        DeepLabv3+~\cite{chen2018encoder}\ &Resnet-18&0.784&0.510&0.424&0.573&0.645&0.897&0.747&0.684&12.3\\
        UNET &Resnet-18&0.827&0.560&0.492&0.641&0.715&0.879&0.805&0.725&14.3\\
        UNET++~\cite{zhou2018unet++} &Resnet-18&0.836&0.571&0.506&0.653&0.727&0.886&0.811&0.734&15.9\\
        LinkNet~\cite{chaurasia2017linknet} &Resnet-18&0.794&0.569&0.484&0.635&0.698&0.865&0.804&0.711&11.7\\
        MaNet~\cite{fan2020ma} &Resnet-18&0.844&\underline{0.587}&0.516&0.663&0.735&\underline{0.906}&0.816&0.755&21.7\\
        AIFN~\cite{li2019attentional} &Resnet-18& 0.799 & 0.541 & 0.486 & 0.645 & 0.719 & 0.845  &0.783 &0.685& 18.3  \\
        Focal-UNET~\cite{jaffari2021novel} &Resnet-18&0.784  &0.577  &0.504  &0.662  &0.724  &0.836 &0.811 &0.700 &18.4   \\
        Pix2pix~\cite{isola2017image} &Resnet-6&0.822  &0.577   &0.509 &0.663  &0.733  &0.872&\underline{0.833}  &0.742  & 10.6   \\ 
        GcGAN~\cite{fu2019geometry}&Resnet-6&0.837  &0.556 &0.501  &0.655  &0.737  & 0.89 &0.795  &0.724 & 13.4   \\
        AFM~\cite{xue2019learning}&UNET&  0.495&0.432 &0.307 &0.457  &0.498  &0.721  &0.684  &0.579  & 44.0  \\ 
        LCNN~\cite{zhou2019end}&Hourglass network&0.541  &0.464  &0.315 &0.498  &0.519  &0.833  &0.717  &0.627& 10.9    \\
        HAWP ~\cite {xue2020holistically} &Hourglass network&0.581  &0.421&0.315  &0.485  &0.532   &0.862  &0.704  &0.633 & 11.6   \\
        PLGAN (ours)\ &Resnet-6&\textbf{0.863}  & 0.577 &\textbf{0.533} & \textbf{0.687} &\textbf{0.769}  & 0.897 & \textbf{0.849} & \textbf{0.787} & 14.9  \\ \hline
    \end{tabular}
\end{adjustbox}
\end{table*}

\section{Experiments}
\label{sec:results}
The experimental results are presented in this section, with comparisons to the state-of-the-art methods.
\subsection{Datasets}

\begin{figure*}[!t]
	\centering

	\hspace{-5mm}\begin{tabular}{p{1.8cm}p{1.8cm}p{1.8cm}p{1.8cm}p{1.8cm}p{1.8cm}p{1.8cm}p{1.8cm}}

	\footnotesize Real&
	\footnotesize GT&
	\footnotesize UNet&
	\footnotesize FPN&
	\footnotesize Pix2Pix&
	\footnotesize HAWP&
		\footnotesize PLGAN-sem&
		\footnotesize PLGAN-HL\\

 	\includegraphics[scale=0.25]{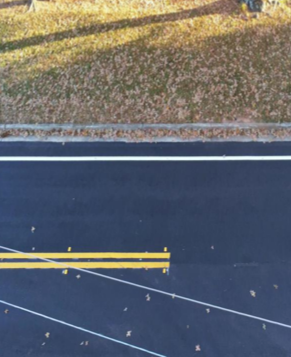}&
 	\includegraphics[scale=0.25]{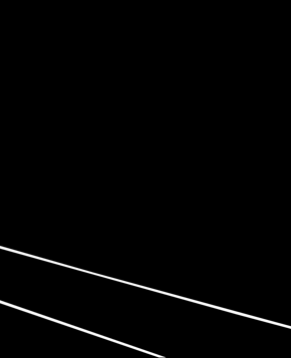}&
 	 	\includegraphics[scale=0.25]{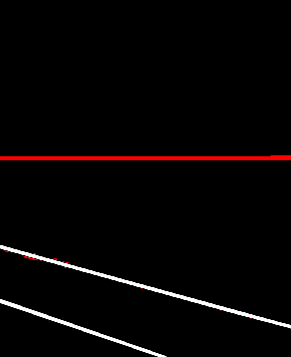}&
 	\includegraphics[scale=0.25]{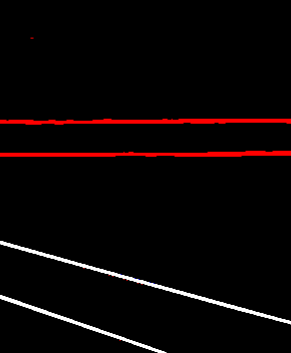}&
 	 \includegraphics[scale=0.25]{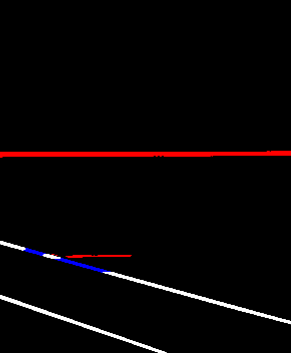}&
 	\includegraphics[scale=0.25]{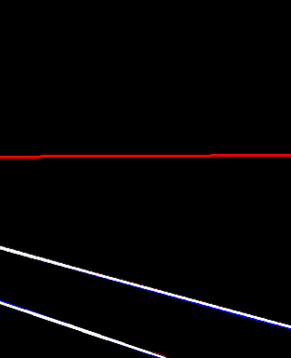}&
 	 	 \includegraphics[scale=0.25]{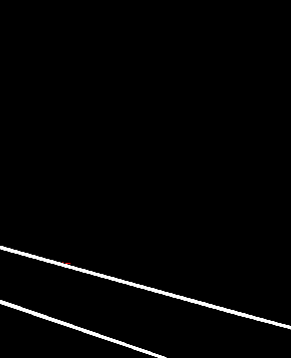}&
 	 	 \includegraphics[scale=0.25]{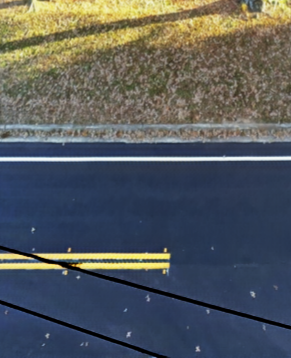}\\
 	\includegraphics[scale=0.25]{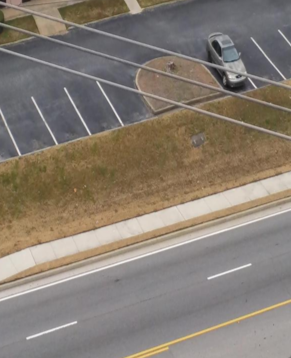}&
 	\includegraphics[scale=0.25]{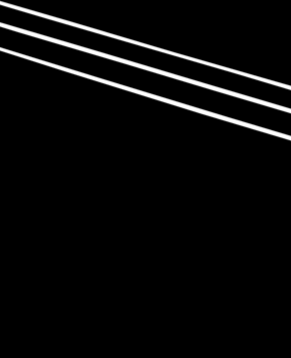}&
 	 	\includegraphics[scale=0.25]{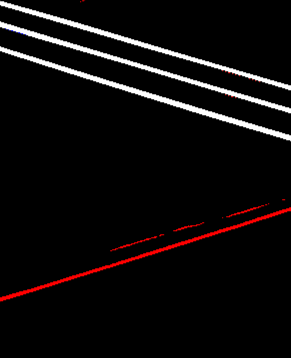}&
 	\includegraphics[scale=0.25]{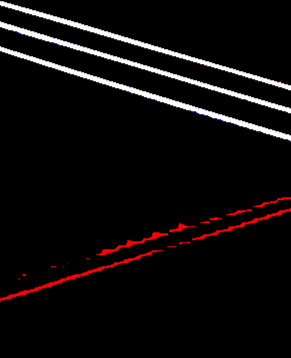}&
 	 \includegraphics[scale=0.25]{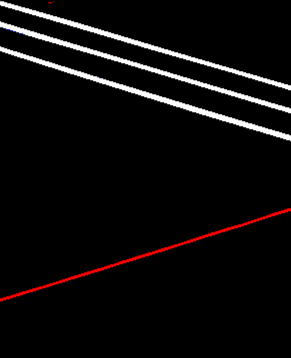}&
 	\includegraphics[scale=0.25]{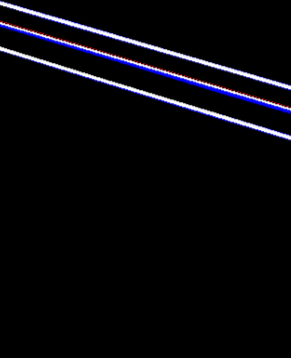}&
 	 	 \includegraphics[scale=0.25]{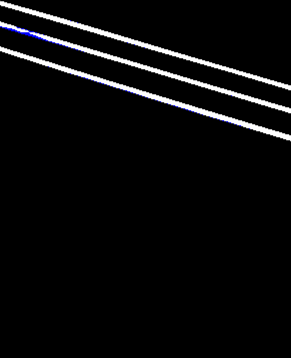}&
 	 	 \includegraphics[scale=0.25]{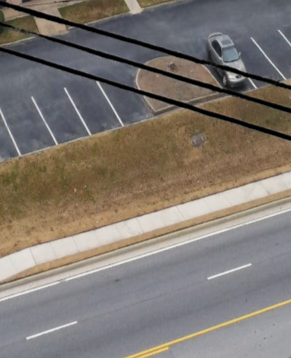}\\
 
 	\includegraphics[scale=0.25]{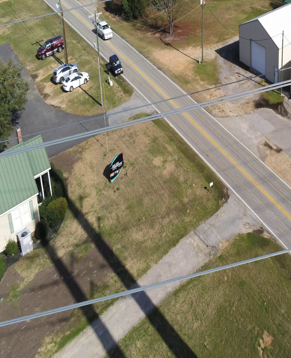}&
 	\includegraphics[scale=0.25]{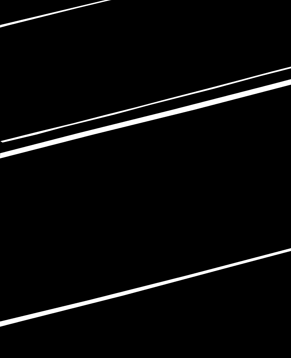}&
 	 	\includegraphics[scale=0.25]{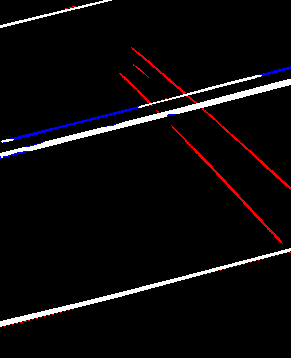}&
 	\includegraphics[scale=0.25]{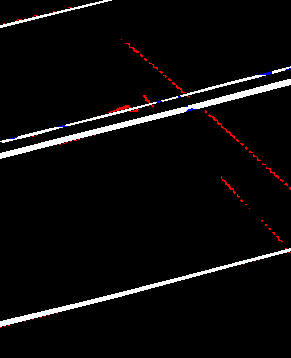}&
 	 \includegraphics[scale=0.25]{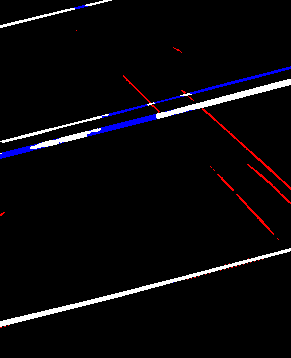}&
 	\includegraphics[scale=0.25]{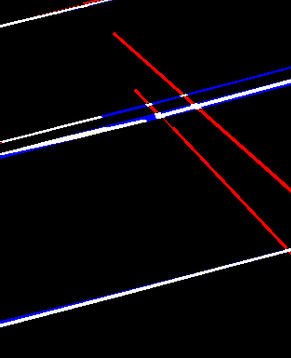}&
 	 	 \includegraphics[scale=0.25]{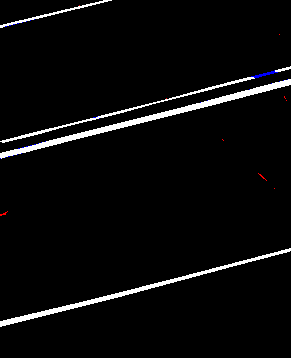}&
 	 	 \includegraphics[scale=0.25]{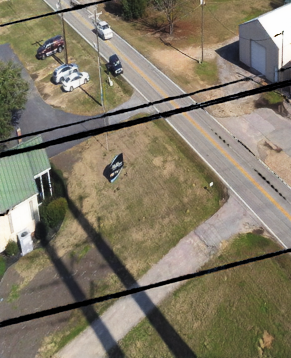}\\

 	\includegraphics[scale=0.25]{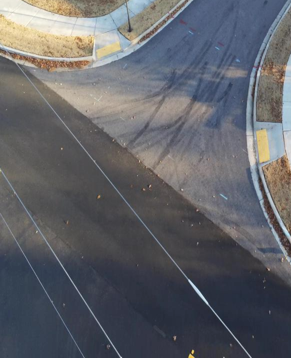}&
 	\includegraphics[scale=0.25]{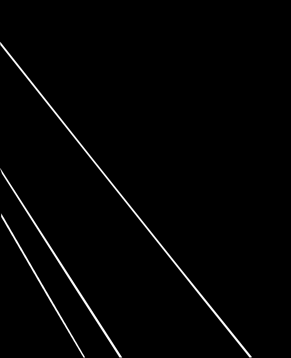}&
 	 \includegraphics[scale=0.25]{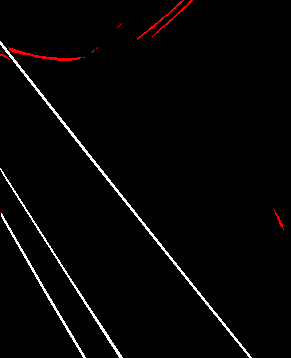}&
 	\includegraphics[scale=0.25]{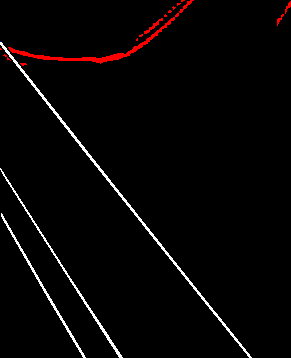}&
 	 \includegraphics[scale=0.25]{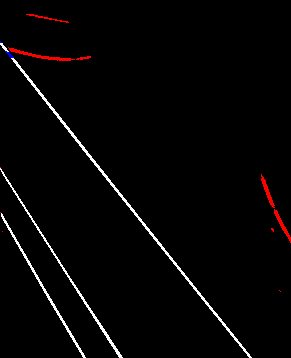}&
 	\includegraphics[scale=0.25]{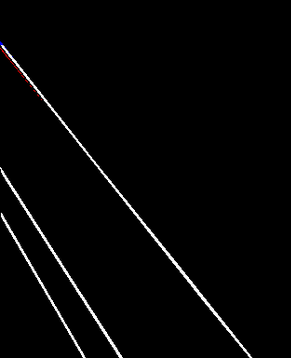}&
 	 	 \includegraphics[scale=0.25]{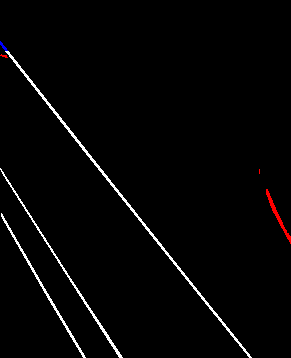}&
 	 	 \includegraphics[scale=0.25]{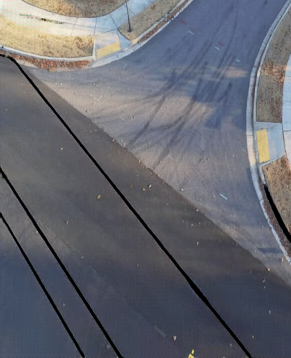}\\


 \includegraphics[width=1.9 cm, height=2.5 cm]{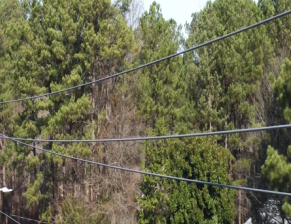}&
 	\includegraphics[width=1.9 cm, height=2.5 cm]{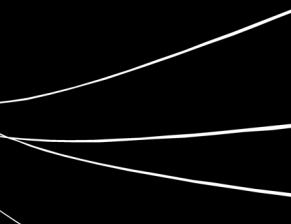}&
 	 	\includegraphics[width=1.9 cm, height=2.5 cm]{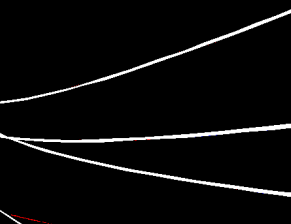}&
 	\includegraphics[width=1.9 cm, height=2.5 cm]{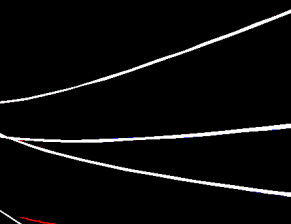}&
 	 \includegraphics[width=1.9 cm, height=2.5 cm]{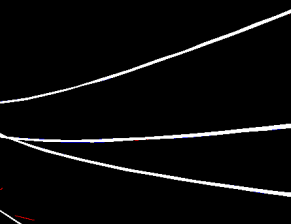}&
 	\includegraphics[width=1.9 cm, height=2.5 cm]{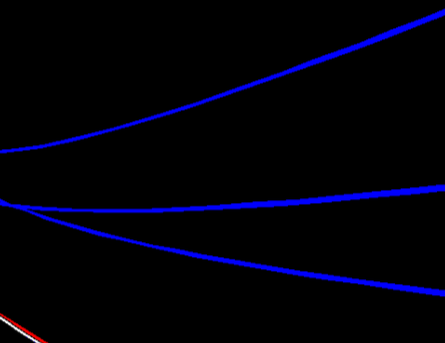}&
 	 	 \includegraphics[width=1.9 cm, height=2.5 cm]{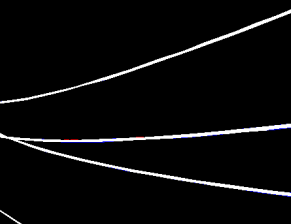}&
 	 	 \includegraphics[width=1.9 cm, height=2.5 cm]{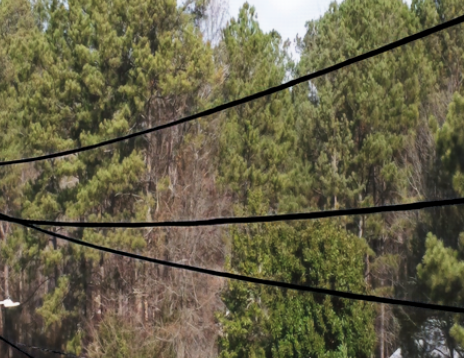}\\
 	\includegraphics[scale=0.25]{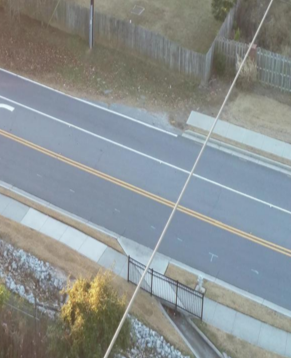}&
 	\includegraphics[scale=0.25]{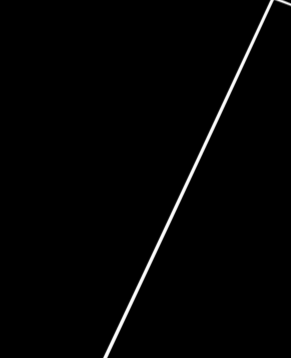}&
 	 	\includegraphics[scale=0.25]{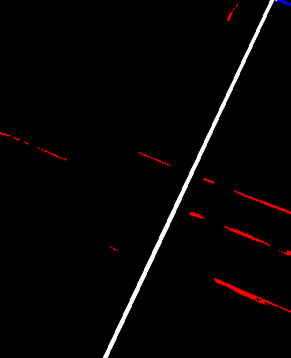}&
 	\includegraphics[scale=0.25]{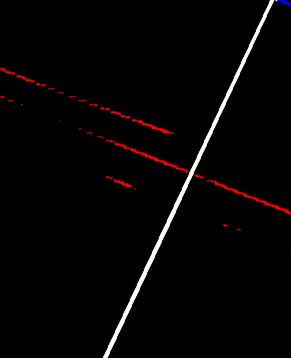}&
 	 \includegraphics[scale=0.25]{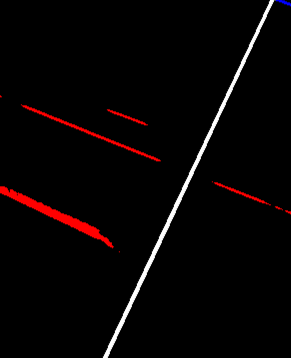}&
 	\includegraphics[scale=0.25]{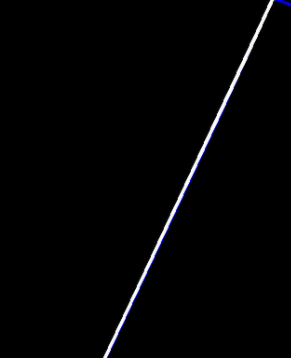}&
 	 	 \includegraphics[scale=0.25]{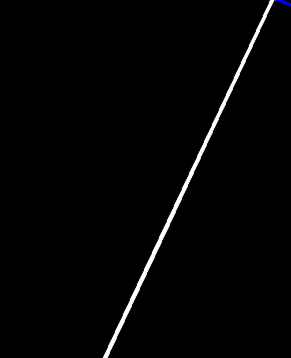}&
 	 	 \includegraphics[scale=0.25]{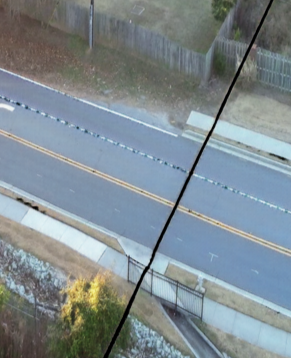}\\

 	\includegraphics[scale=0.25]{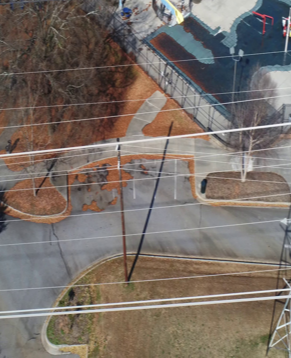}&
 	\includegraphics[scale=0.25]{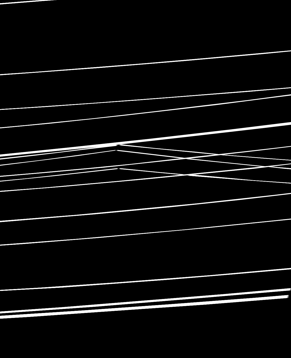}&
 	 	\includegraphics[scale=0.25]{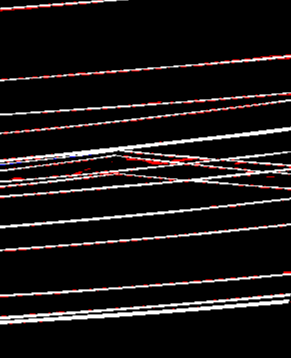}&
 	\includegraphics[scale=0.25]{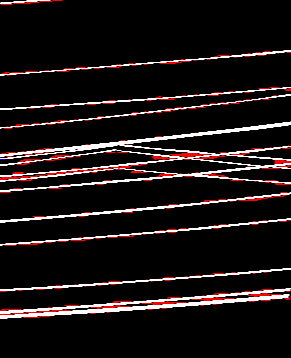}&
 	 \includegraphics[scale=0.25]{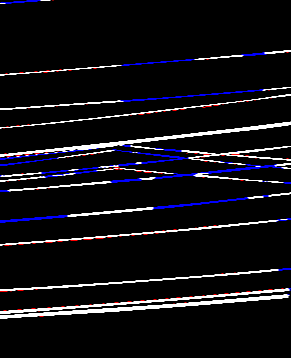}&
 	\includegraphics[scale=0.25]{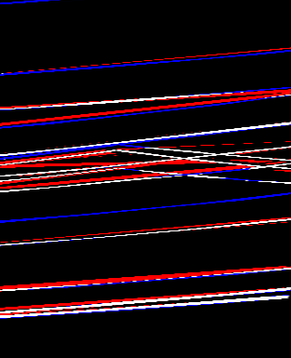}&
 	 	 \includegraphics[scale=0.25]{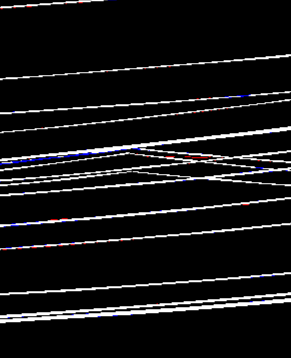}&
 	 	 \includegraphics[scale=0.25]{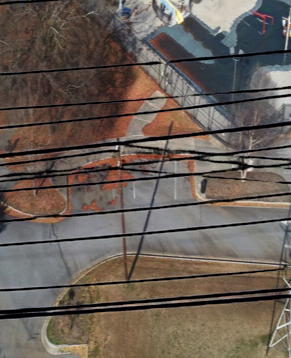}\\
	\end{tabular}
	\caption{Sample PL segmentation results produced by the proposed PLGAN and comparison methods. The blue and red colors indicate the missing and false predication, respectively. Appearing both colors for the same line means that this line has small curve which can not detected correctly. Two pixels relaxation are used for all models to make the visualization more clear.  }
	\vspace{-0.5mm}
	\label{fig:baselines}
\end{figure*}

TTPLA \cite{abdelfattah2020ttpla} is a public dataset that contains aerial images for PLs from different zoom levels and view angles, collected at different time and locations with different backgrounds. TTPLA dataset contains 8,083 instances of PLs, which take only 154M pixels, 1.68\% of the total number of pixels in this dataset~\cite{abdelfattah2020ttpla}. This dataset contains about 1,100 images. We used 905 training images, augmented by vertical/horizontal flipping, and 217 images for test set. Each instance of PL is carefully annotated by a polygon using LabelME~\cite{russell2008labelme}.
TTPLA also provides polygonal annotations of all the transmission present in each image, and an instance of PL is usually considered to be ended when it enters the annotated polygon of transmission tower, as shown the second column of Fig~\ref{fig:baselines}. 
Since there are few public PL datasets available, we also considered Massachusetts Roads dataset~\cite{mnih2013machine} instead to further evaluate the performance of PLGAN.  This dataset is used for road segmentation, which consists of 1,108 training and 49 test images, including both urban and rural neighborhoods with pixel level annotations.

\subsection{Implementation Details}
The proposed PLGAN is implemented using PyTorch and trained with a single NVIDIA Tesla V100 GPU with 16GB.  
The weights of all sub-nets are initialized based on normal distribution using Xavier method with zero mean and gain $0.02$. They are jointly optimised using Adam with the first and the second momentum setting to $0.5$ and $0.999$, respectively.
The entire model is trained for 200 epochs with the image size of $512 \times 512$. The learning rate starts with $1 \times 10^{-4}$ for the first 100 epochs and decays to zero during the second 100 epochs. All models are trained from scratch. The ground-truth of the PL-highlighted images are obtained by simply setting the intensity of the PL pixels in the images to zero. PLGAN uses ResNet as a backbone, following CyclicGAN, GcGAN, and Pix2Pix GAN, and the training starts with Gaussian distribution (mean 0 and std 0.02).
\subsection {Evaluation Metrics}
We adopt totally eight metrics to evaluate the detection performance of our model. Precision, recall, and intersection-over-union (IoU) are the widely used metric in semantic segmentation~\cite{wang2019recurrent}. Also, we consider $F$ score as an evaluation metric that is the harmonic mean of average precision and average recall.  It is defined as $ F_{\beta}= \frac{(1+{\beta}^2) Precision \times Recall}{{\beta}^2Precision + Recall}$, where we assign $\beta$ with two values: $\beta = 1$ following \cite{mosinska2018beyond} and $\beta = 0.3$ to emphasize more precision over recall which follows \cite{mei2020don}.  Furthermore, we investigate the completeness (comp.), correctness (corr.), and quality as the evaluation metrics, following the previous studies on thin-object detection~\cite{mosinska2018beyond, qin2019basnet,wiedemann1998empirical,zou2018deepcrack, sironi2014multiscale}. 
Under these metrics, the definition of true positives can be extended to the case that allows the predicted pixel to shift a certain distance from its ground truth. Correctness and completeness represent the extended precision and recall, respectively, while quality $= \frac{comp. \times corr.}{comp. -comp. \times corr. + corr.}$. In our experiments, we allow the shift to be 2 pixels under these three evaluation metrics, following~\cite{mosinska2018beyond, sironi2014multiscale}. 

\subsection{Comparison with Existing Methods on TTPLA Dataset}
We compare the performance of PLGAN on TTPLA with a number of existing methods that can be grouped into three different categories. ($i$) Semantic image segmentation models:   LinkNet~\cite{chaurasia2017linknet}, UNet++~\cite{zhou2018unet++}, FPN~\cite{lin2017feature}, DeepLabv3+~\cite{chen2018encoder}, UNET~\cite{ronneberger2015u}, MaNet~\cite{fan2020ma}, AIFN \cite{li2019attentional} and Focal-UNet \cite{jaffari2021novel} as reported in Table \ref{tab:results-basics}; 
($ii$) GAN-based architectures: Pix2pix \cite{isola2017image} and GcGAN \cite{fu2019geometry} based on backbone 6 residual blocks (ResNet-6).  GANs are evaluated based on the semantic images generated by assigning one to the pixels belonging to PLs and zero otherwise; 
($iii$) Line segment detectors:  AFM~\cite{xue2019learning}, LCNN~\cite{zhou2019end}, and HAWP~\cite{xue2020holistically}.  AFM uses UNET as the backbone while the other two rely on stacked Hourglass network~\cite{newell2016stacked} as the backbone. Since the line segment detectors require different type of annotation for their ground truth depending on start and end points for each line which is not compatible with our setting, we extend it to our problem, to compare with them, by prepare line segment annotation of PL on all the images in TTPLA, by following the general annotation pipeline in~\cite{xue2019learning} on the original polygonal PL annotations. 

Tables~\ref{tab:results-basics} shows the quantitative results of the proposed PLGAN and all above comparison methods on the test set of TTPLA dataset. 
Figure~\ref{fig:baselines} shows the segmentation results of sample images from both the proposed PLGAN and the comparison~methods. 

\smallskip
\noindent
\textbf{Comparison with Deep Semantic Segmentation Models.}
It is shown in Table~\ref{tab:results-basics} that PLGAN outperforms most of the baselines.  We found that, compared with PLGAN, those baseline models produce more false positives in PL segmentation. For instance, UNET and FPN (columns 3 and 4 in Figure~\ref{fig:baselines}) mis-classified many non-PL structures, such as sidewalks and lanes, as part of PLs.  This observation can be interpreted from two aspects. First, most of these models are built upon the encoder-decoder structures, while the decoders fail to appropriately augment the complex background information when making pixel-wise predictions from the low resolution feature maps generated by the encoder~\cite{tian2019decoders}. Second, the networks are trained based on the softmax cross-entropy loss and ignore the interconnections between pixels as discussed in context~\cite{ke2018adaptive,zhao2019region}. Therefore, it is hard to preserve global consistency~\cite{samson2019bet}. Even though Focal-UNET \cite{jaffari2021novel} uses focal loss function instead of BCE loss function for addressing the class imbalance in PL segmentation, it still suffers from the same limitation by not capturing the relation between pixels. We also notice that, although UNET++ and MaNet using ResNet-34 outperform PLGAN in recall and correctness, respectively, it is at a cost of many more parameters than PLGAN. 

\smallskip
\noindent
\textbf{Comparison with GANs.}
As shown in column 5 of Figure~\ref{fig:baselines}, the detection from pix2pix GAN, the direct use of GANs for semantic segmentation reduces the performance by missing many PL pixels and generating false positives, resulting in many gaps along the segmented PLs. 
This is also reflected in the quantitative results shown in Table~\ref{tab:results-basics}.
As discussed in Related Work Section, this is the inherited limitation when  generating/discriminating the semantic images directly: the discriminator pushes the generator to produce semantic images with sharp zeroes/ones and leave a permanent possibility for the discriminator to examine the small, but always existing, value gap between the distributions of true labels and the predictions~\cite{samson2019bet}, which may hurt the performance of adversarial training. As shown in Table~\ref{tab:results-basics}, instead of directly using GAN to generate semantic images, the proposed PLGAN embeds features from GAN to an semantic segmentation network and can achieve much higher quality in PL segmentation. More Figures for GAN results are presented in supplementary material. 

\smallskip
\noindent
\textbf{Comparison with Line Segment Detector.} As shown in Figure~\ref{fig:baselines} (column 6), most of line detectors can capture many PLs with very clean segmentation. 
This is totally reasonable since PLs are very-thin line structures and line detectors fully take advance of this geometry prior to ensure the global consistency in PL segmentation. However, in using deep neural networks to boost the capability of line segment detection, most of line detectors conducts spatial-region partitioning for network computation and feature representation. This inherently reduces the spatial resolution of features and may cause dislocation between the segmented PLs and their corresponding GTs. As a result, group of lines can be missing in Figure~\ref{fig:baselines}. In  addition,  the line segment detectors, cannot handle the curved power lines as shown in the image at column 6 and row 5). Therefore, while most of line segment detectors produce quite clean PL segmentation in some cases, its quality is still much lower than our PLGAN, as shown in Table~\ref{tab:results-basics}. 

\subsection{Comparison on Massachusetts Roads Dataset} 
Due to the lack of public PL datasets, we evaluate PLGAN on Massachusetts roads dataset for road extraction, which has the same nature as thin objects.  We first follow the experiment setting in~\cite{wang2019recurrent} and evaluate PLGAN using precision, recall, IoU, and $F_1$ score. We compare the performance of PLGAN with Rec-Middle~\cite{poudel2016recurrent}, Rec-Last~\cite{valipour2017recurrent}, ICNet~\cite{zhao2018icnet}, Rec-Simple~\cite{mosinska2018beyond}, and DRU~\cite{wang2019recurrent}.  The results are reported in Table~\ref{tab:road_seg}.
Then we follow the experiment setup in~\cite {mosinska2018beyond} to evaluate completeness, correctness, and quality of PLGAN.  We compare our performance with Reg-AC~\cite{sironi2015multiscale}, MNIH~\cite{mnih2013machine}, and Rec-Simple~\cite{mosinska2018beyond}. The results are reported in Table \ref{tab:road_topology}. 
In addition, we provide the results for Deeplab V3++, LinkNet, MaNet, and Unet++ using Resnet-34 as the backbone in both Tables~\ref{tab:road_seg} and~\ref{tab:road_topology}.

\begin{figure*}[!t]
	\centering
	\hspace{-5mm}\begin{tabular}{p{2.2cm}p{2.2cm}p{2.2cm}p{2.2cm}p{2.2cm}p{2.2cm}}

	\footnotesize Real&
	\footnotesize GT&
	\footnotesize LinkNet&
		\footnotesize MaNet&
	\footnotesize PLGAN-sem&
	\footnotesize PLGAN-HL\\
 	\includegraphics[scale=0.17]{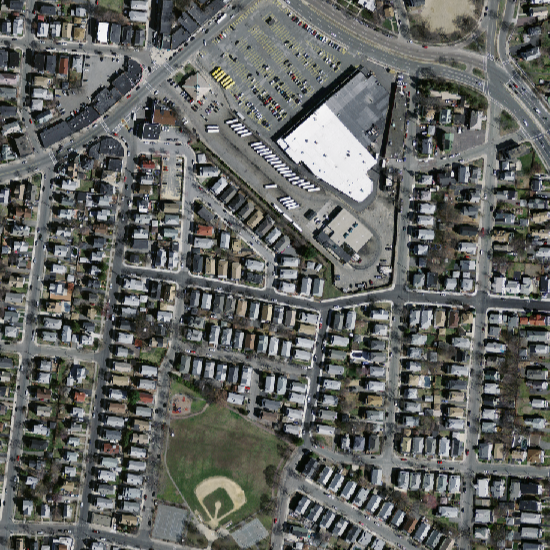}&
 	\includegraphics[scale=0.17]{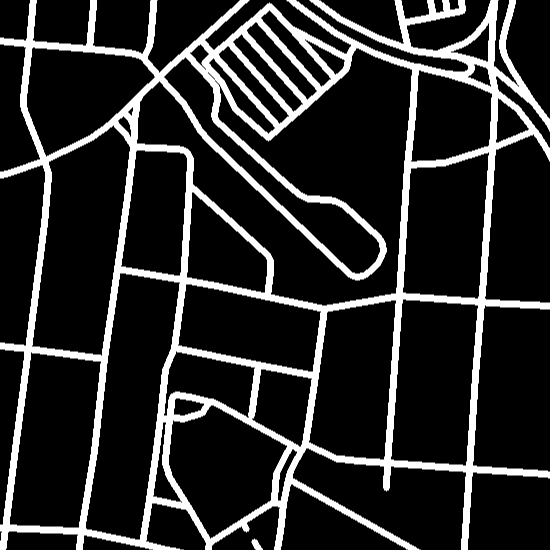}&
 	\includegraphics[scale=0.17]{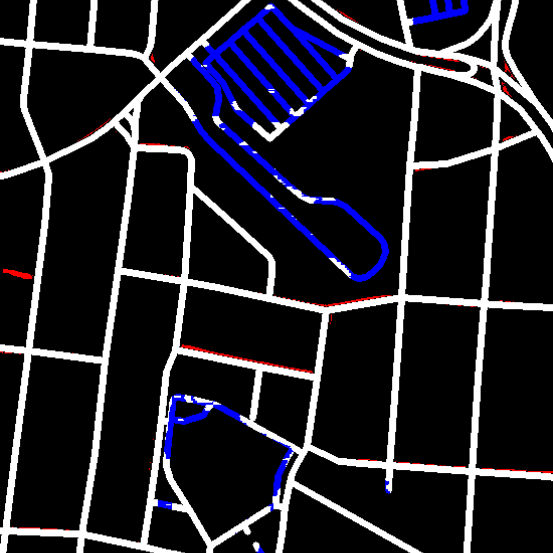}&
 	\includegraphics[scale=0.17]{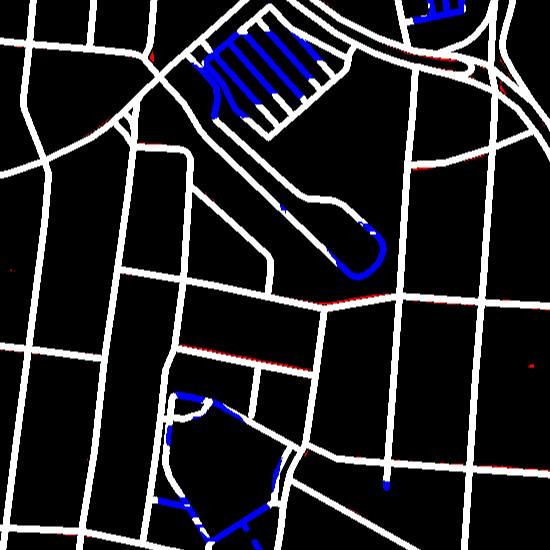}&
 	\includegraphics[scale=0.17]{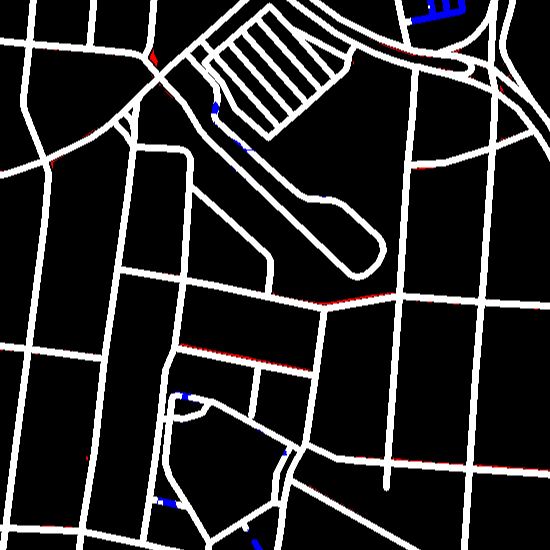}&
 	 \includegraphics[scale=0.17]{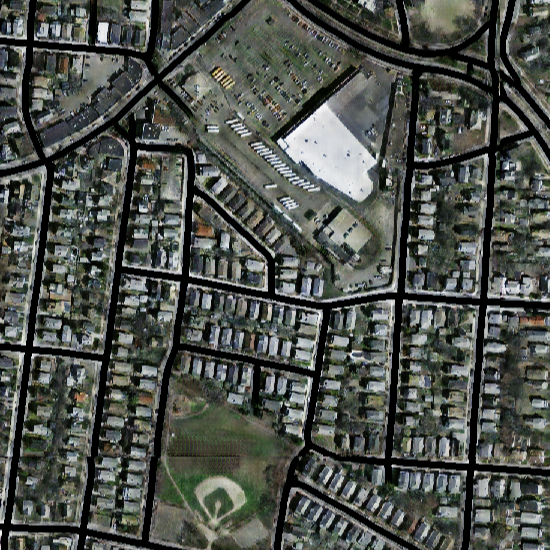}\\
 	\includegraphics[scale=0.17]{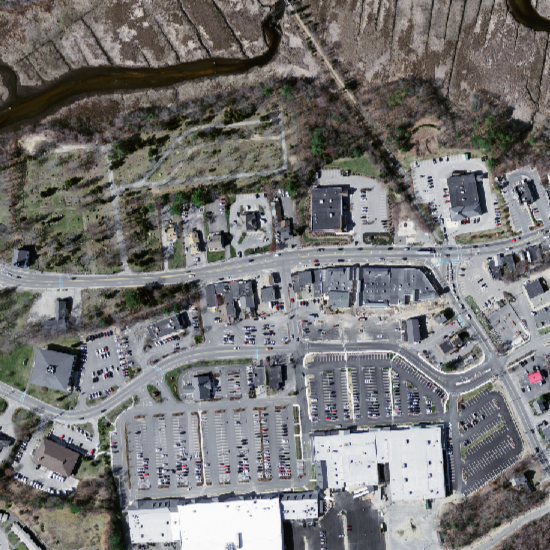}&
 	\includegraphics[scale=0.17]{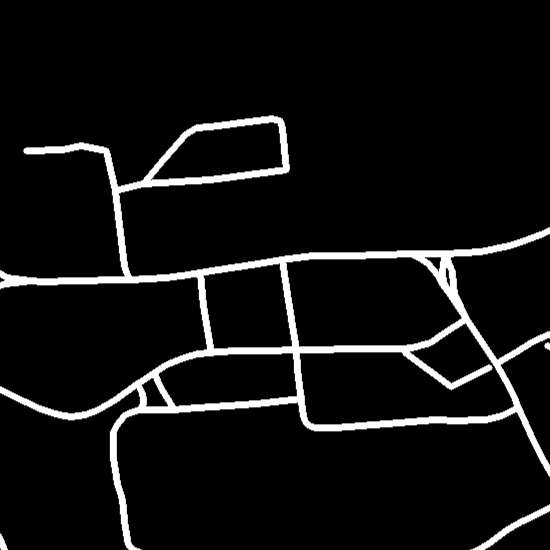}&
 	\includegraphics[scale=0.17]{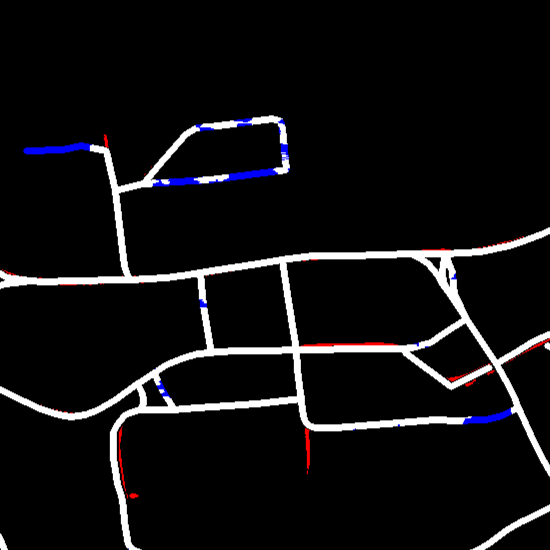}&
 	\includegraphics[scale=0.17]{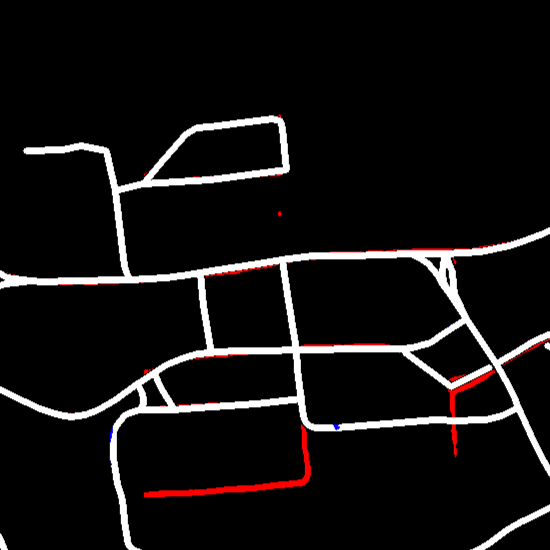}&
 	\includegraphics[scale=0.17]{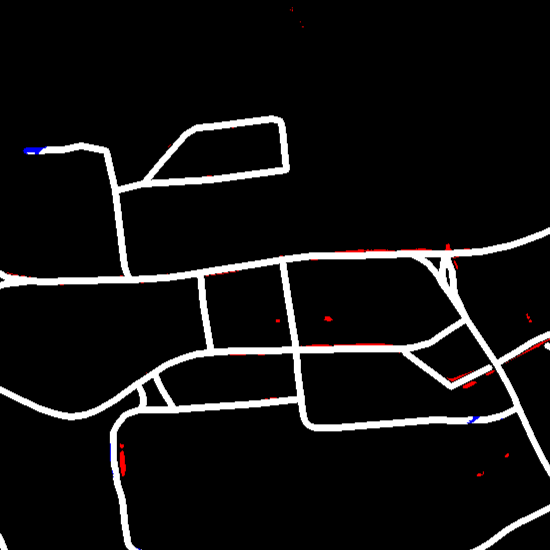}&
 	 \includegraphics[scale=0.17]{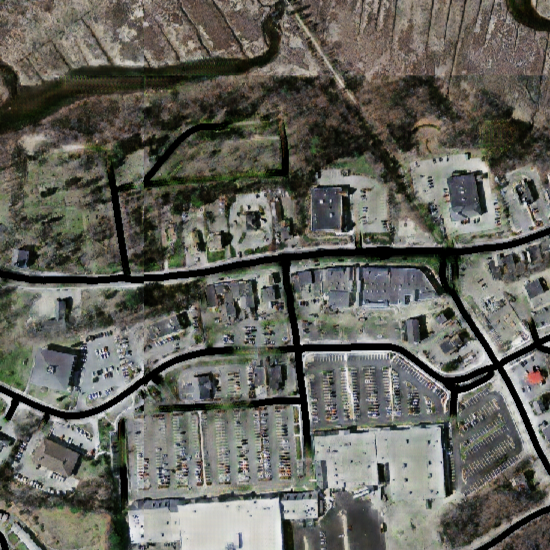}\\
 	\includegraphics[scale=0.17]{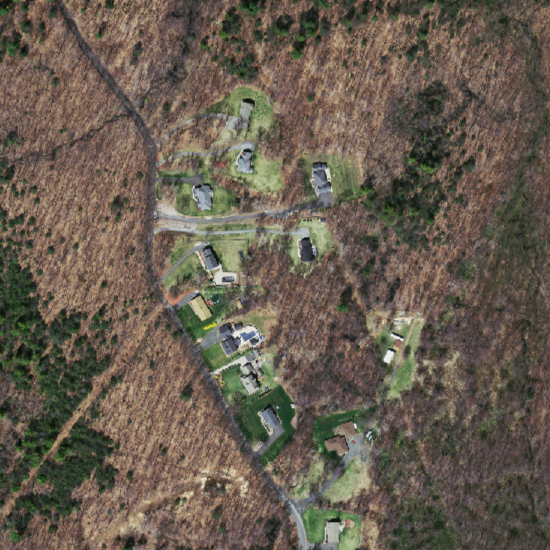}&
 	\includegraphics[scale=0.17]{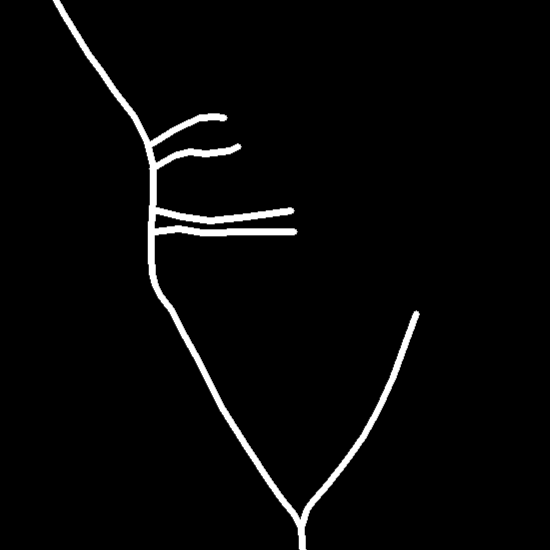}&
 	\includegraphics[scale=0.17]{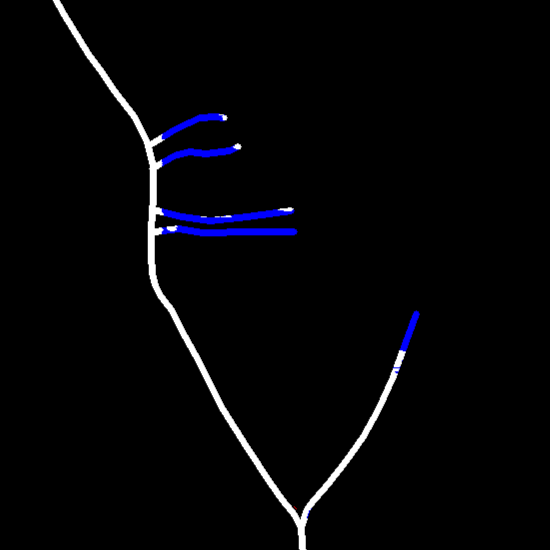}&
 	\includegraphics[scale=0.17]{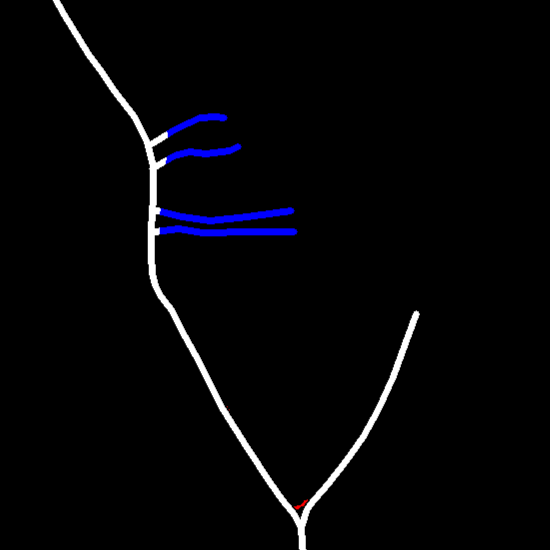}&
 	\includegraphics[scale=0.17]{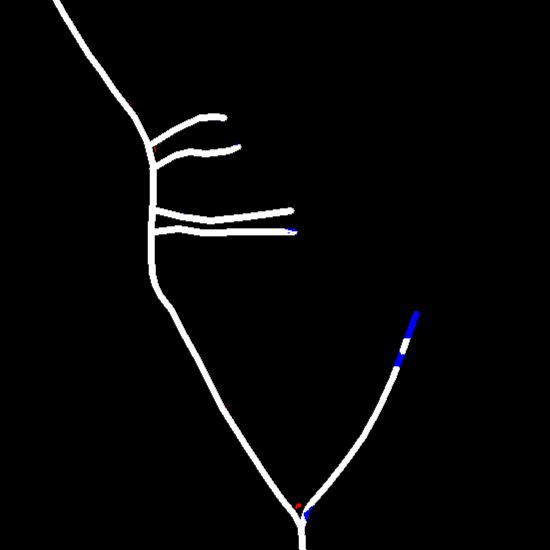}&
 	 \includegraphics[scale=0.17]{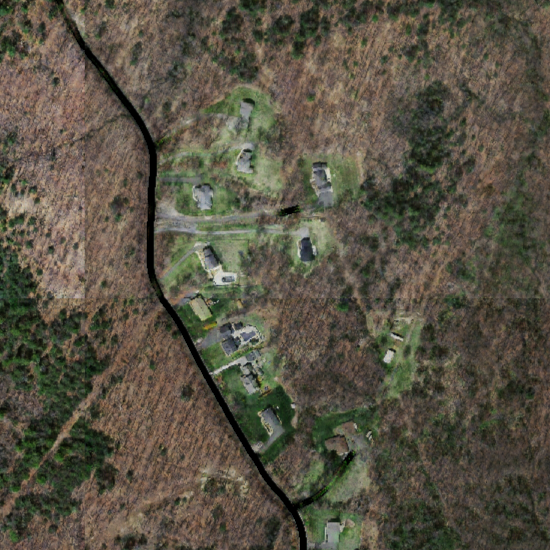}\\
 	\includegraphics[scale=0.17]{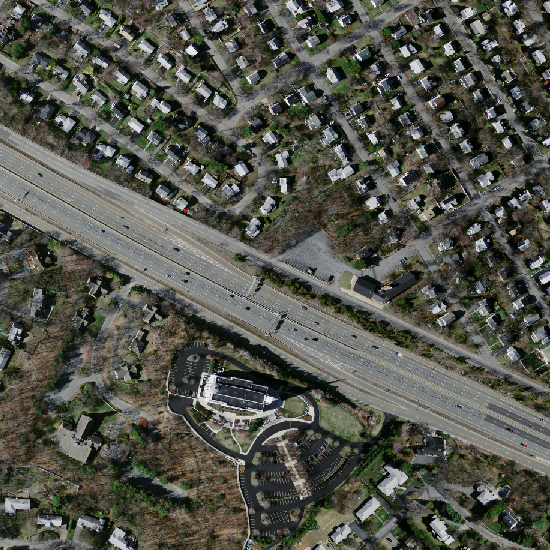}&
 	\includegraphics[scale=0.17]{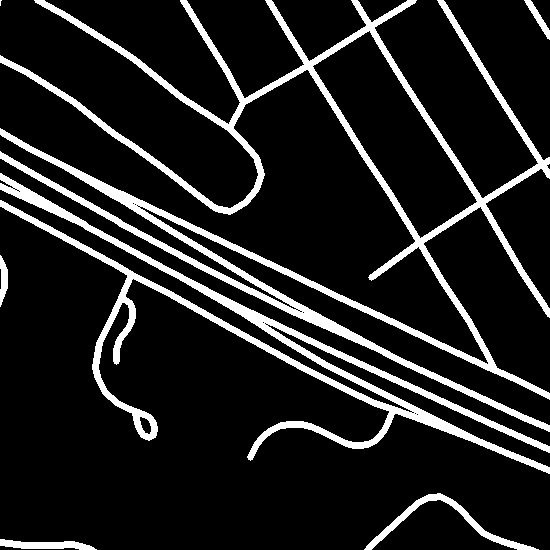}&
 	\includegraphics[scale=0.17]{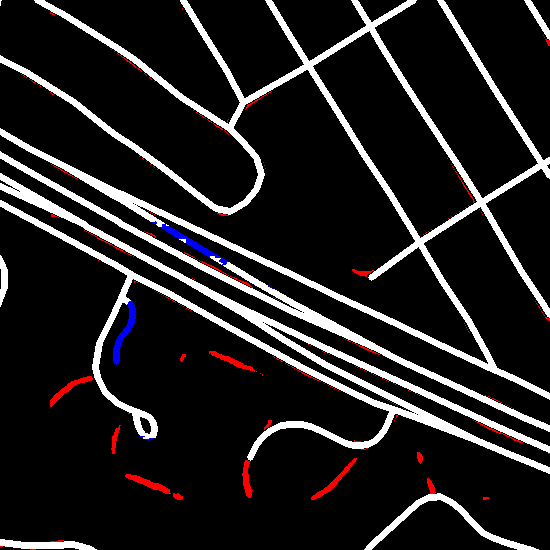}&
 	\includegraphics[scale=0.17]{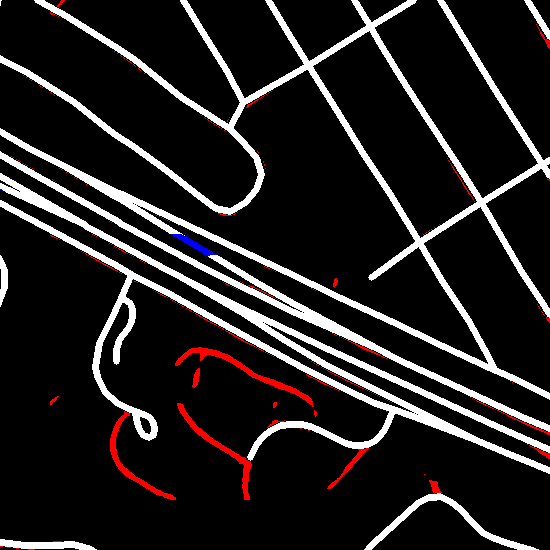}&
 	\includegraphics[scale=0.17]{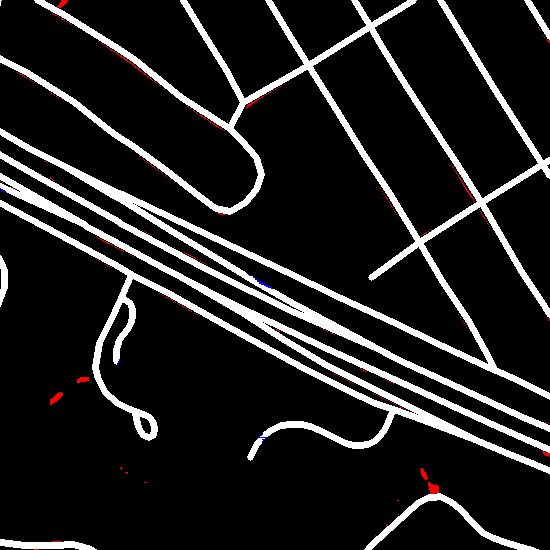}&
 	 \includegraphics[scale=0.17]{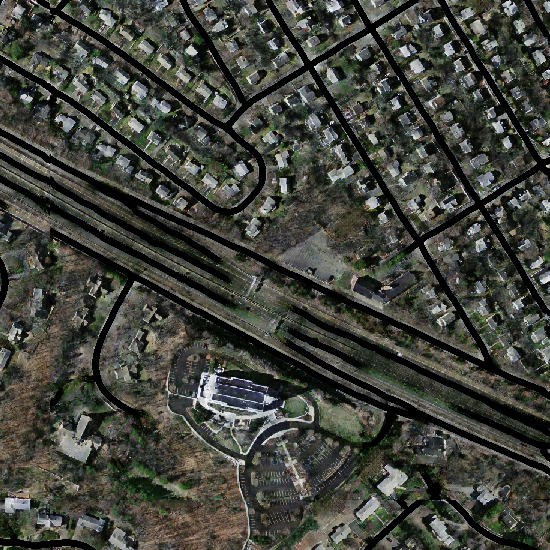}\\
	\end{tabular}
	\caption{Road extraction by our proposed PLGAN. The blue and red colors indicate the missing and false predication, respectively. Two pixels relaxation are used for all models to make the visualization more clear.}
	\vspace{-2mm}
	\label{fig:road}
\end{figure*}

It can be found from both tables that PLGAN outperforms the state-of-the-art methods under most evaluation metrics.
Our PLGAN achieves highest precision, IoU, and $F_1$ as reported in Table~\ref{tab:road_seg} and the best completeness and quality in Table~\ref{tab:road_topology}.  It is worth mentioning that UNet++ model in Table~\ref{tab:road_seg} and MaNet in Table~\ref{tab:road_topology} achieve the second best $F_1$ and quality, respectively. This is mainly because UNet++ and MaNet use a significantly larger number of parameters (26.1 and 31.8, respectively) than PLGAN (14.9). Some segmentation testing samples are shown in Figure~\ref{fig:road}.

\begin{table}[!t]
\center
\caption{Comparison on Massachusetts roads dataset by Precision, Recall, IoU, and $F_1$ score. Bold represents the highest results and underline represents the second-best.}
\begin{adjustbox}{width=0.48\textwidth}  
\begin{tabular}{l|c|c|c|c}
\hline
Models &Precision &Recall & IoU &$F_1$ \\
\hline
Rec-Middle~\cite{poudel2016recurrent}&0.518&0.767&0.494&0.574\\
Rec-Last~\cite{valipour2017recurrent}&0.551&0.786&0.526&0.648\\
ICNet~\cite{zhao2018icnet} &0.500&0.626 &0.476 &0.656\\ 
Rec-Simple~\cite{mosinska2018beyond}&0.559&\underline{0.802}&0.534&0.659\\
Linknet~\cite{chaurasia2017linknet}&0.785&0.661&0.523&0.676\\
Deeplab V3+~\cite{chen2018encoder}   &0.773&0.667  &0.525 &0.678\\
MaNet~\cite{fan2020ma}&0.789&0.681&0.539&0.689\\
DRU~\cite{wang2019recurrent}   &0.583 &\textbf{0.865} &\underline{0.560}&0.691\\
UNet++~\cite{zhou2018unet++}&\underline{0.807}&0.655&0.540&\underline{0.694}\\
\textbf{PLGAN (Ours)} &\textbf{0.813}&0.691&\textbf{0.571}&\textbf{0.721}\\
\hline

\end{tabular}
\vspace{0mm}
\end{adjustbox}

\label{tab:road_seg}
\end{table}
\begin{table}[!t]
\center
\caption{Comparison on Massachusetts roads dataset by Completeness, correctness, and quality. Bold represents the highest results and underline represents the second-best.}
\begin{adjustbox}{width=0.42\textwidth}  
\begin{tabular}{l|c|c|c}
\hline
Models &Corr. &Comp. & Quality \\
\hline
Reg-AC~\cite{sironi2015multiscale} &0.254  &0.348 &0.172\\
MNIH~\cite{mnih2013machine} &  0.531  &0.752&0.452\\ 
Rec-Simple~\cite{mosinska2018beyond}   &0.774 &0.806 &0.652\\

Linknet~\cite{chaurasia2017linknet}&0.919&0.820&0.757\\
Deeplab V3+~\cite{chen2018encoder} &0.914&0.822&0.756\\
MaNet~\cite{fan2020ma}&0.922&\underline{0.828}&\underline{0.766}\\
UNet++~\cite{zhou2018unet++}&\textbf{0.943}&0.804&0.763\\
\textbf{PLGAN (Ours)} &\underline{0.937}&\textbf{0.833}&\textbf{0.788}\\
\hline
\end{tabular}
\vspace{0mm}
\end{adjustbox}

\label{tab:road_topology}
\vspace{-0.5cm}
\end{table}
\subsection{Ablation Study} 
We evaluate the performance of several variants of the proposed PLGAN to justify the usefulness of its different components. The ablation study is performed on TTPLA dataset. We start with the baseline variant which simply uses the PL-aware generator ($G$) with the adversarial loss function to directly generate the semantic segmentation images.  The results are listed in row 1 of Table~\ref{tab:ablation}. The second variant of PLGAN applies our idea of using the semantic decoder ($S$) to produce the semantic images and using the PL-aware generator to generate PL-highlighted images (row 2 in Table~\ref{tab:ablation}). Then we add the hough transform loss function ($\mathcal{HT}$) and the results are listed in row 3.  Finally,  the geometry loss function is applied on top of the previous variant (row 4 in Table~\ref{tab:ablation}).

\begin{figure}[!t]
	\centering
	\hspace{-5mm}\begin{tabular}{p{1.1cm}p{1.1cm}p{1.1cm}p{1.1cm}p{1.1cm}p{1.1cm}}

	\footnotesize Real&
	\footnotesize GT&
	\footnotesize G&
		\footnotesize G+S&
	\footnotesize G+S+HT&
	\footnotesize G+S+HT+geo\\
 	\includegraphics[scale=0.20]{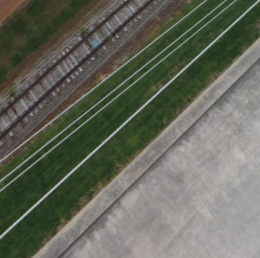}&
 	\includegraphics[scale=0.20]{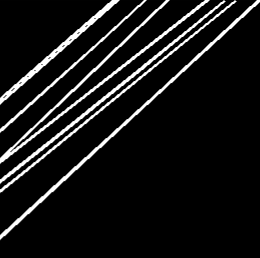}&
 	\includegraphics[scale=0.20]{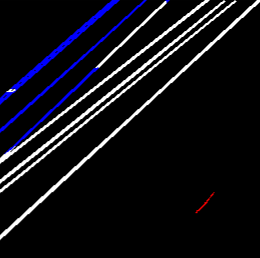}&
 	\includegraphics[scale=0.20]{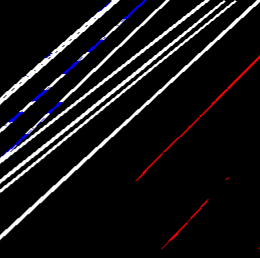}&
 	\includegraphics[scale=0.20]{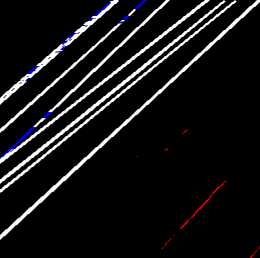}&
 	 \includegraphics[scale=0.20]{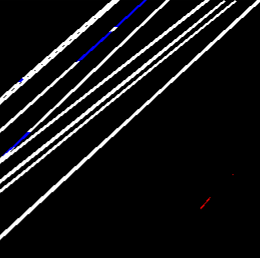}\\
 	\includegraphics[scale=0.20]{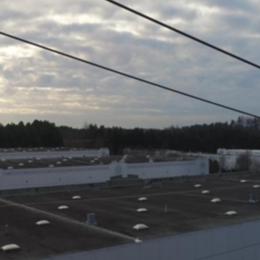}&
 	\includegraphics[scale=0.20]{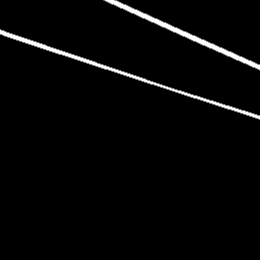}&
 	\includegraphics[scale=0.20]{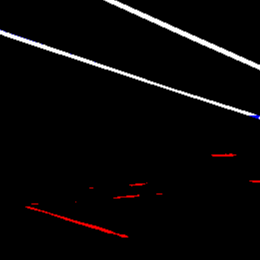}&
 	\includegraphics[scale=0.20]{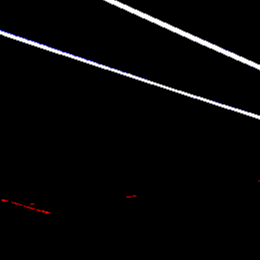}&
 	\includegraphics[scale=0.20]{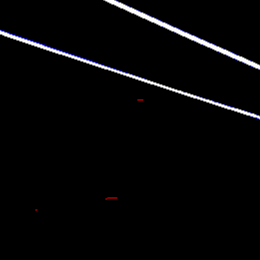}&
 	 \includegraphics[scale=0.20]{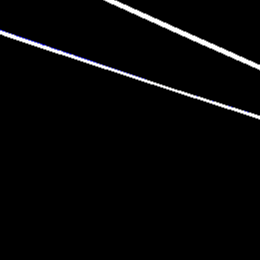}\\
 	\includegraphics[scale=0.20]{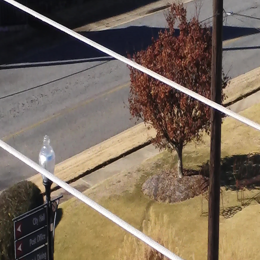}&
 	\includegraphics[scale=0.20]{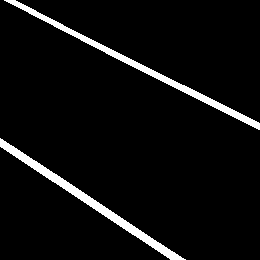}&
 	\includegraphics[scale=0.20]{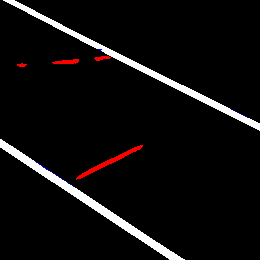}&
 	\includegraphics[scale=0.20]{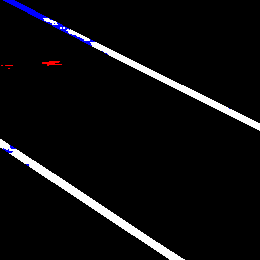}&
 	\includegraphics[scale=0.20]{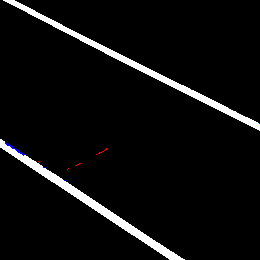}&
 	 \includegraphics[scale=0.20]{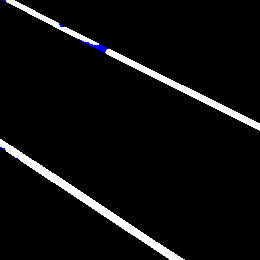}\\
	\end{tabular}
	\caption{Ablation study for different variant of PLGAN. The blue and red colors indicate the missing and false predication, respectively. }
	\label{fig:ablation}
\end{figure}

As shown in Table~\ref{tab:ablation} and Figure~\ref{fig:ablation}, we notice that applying the PL-highlighted images helps the generator to build the embedding vector as an input to semantic decoder. Therefore, the performance across all metrics is improved in row 2.  Also, we observe that $\mathcal{L}_{ht}$ slightly improves the precision and IoU of PLGAN (row3) and $\mathcal{L}_{geo}$ enhances the recall and $F_1$ (row 4).  Based on these observations, our contributions are found complementary and the experimental results validate the importance of building end-to-end trainable models.
\section{Conclusions}
\label{sec:con}
This paper proposes a new GAN framework, PLGAN, for power line segmentation in aerial images. PLGAN takes advantage of adversarial training and captures the context, geometry, and appearance information for prediction.  In PLGAN, the generated PL-highlighted images are used in the discriminator that forces PLGAN to highlight the PL areas in the images. By learning the joint representation in the shared latent space from the PL-highlighted image and the semantic image, PLGAN can generate more accurate semantic images, compared with the state-of-the-art approaches, which is verified through extensive experiments. In fact, PLGAN has the potential to be extended to general models for segmentation that requires global consistency (not limited to lines), which will be investigated in our future work.

\begin{table}[!t]
\center
\caption{Quantitative ablation study of PLGAN based on PL-highlighted generation images.}
\begin{adjustbox}{width=0.48\textwidth, height={1.0cm}}  
\begin{tabular}{c|c|c|c||c|c|c|c|c}
\hline
$G$ & $S$ &$\mathcal{HT}$ &$geo$ 
&Precision &Recall  &IoU&$F_1$& $F_{\beta}$\\
\hline
\checkmark &&&&0.822&0.561&0.509&0.663&0.733\\
\checkmark &\checkmark   & &&0.861&0.565&0.520&0.677&0.762\\
\checkmark&\checkmark&\checkmark&&0.865&0.560&0.524&0.679&0.768\\
\checkmark&\checkmark&\checkmark&\checkmark&0.864&0.577&0.533&0.687&0.770\\
\hline
\end{tabular}
\end{adjustbox}

\label{tab:ablation}

\end{table}

\bibliographystyle{IEEEtran}
\bibliography{3_paper}

\end{document}